\AtBeginDocument{%
  \setlength\abovedisplayskip{0pt}
  \setlength\belowdisplayskip{0pt}}
\documentclass[twoside,leqno,twocolumn]{article}
\usepackage{ltexpprt}
\usepackage{multicol}
\usepackage{graphicx}
\usepackage{textcomp}
\usepackage{subfigure}
\usepackage{array}
\usepackage{amsmath}

\usepackage{enumitem}
\newenvironment{tightitemize}
   {\begin{itemize}
   \setlength{\parskip}{0pt}}
   {\end{itemize}}

\begin{document}
\begin{multicols}{2}

\title{Exploring Temporal Relations for Event Tracking Using Hierarchical Dirichlet Trees}
\author{Rumeng Li\thanks{Peking University, Beijing 100871, China} \\
\and
Tao Wang\thanks{Wuhan University, Wuhan 430072, China}\\
\and
Xun Wang\thanks{Peking University, Beijing 100871, China}
}
\date{}
\maketitle
\end{multicols}

\begin{abstract}
Timeline Generation, through generating news timelines from the massive data of news corpus, aims at providing readers with summaries about the evolvement of an event. It is a new challenge of summarization that combines salience ranking with novelty detection. For a long-term public event, the main topic usually includes many different sub-topics at varying epochs, which also has its own evolving patterns. Existing approaches fail to utilize such hierarchical topic structure involved in the news corpus for timeline generation . In this paper, we develop a novel time-dependent Hierarchical Dirichlet Tree Model (tHDT) for timeline generation. Our model can aptly detect different levels of topic information in corpus and the structure is further used for sentence selection. Based on the topic distribution mined from tHDT, sentences are selected through an overall consideration of relevance, coherence and coverage. We develop experimental systems to compare different rival algorithms on 8 long-term events of public concern. The performance comparison demonstrates the effectiveness of our proposed model in terms of ROUGE metrics.
\end{abstract}




\section{Introduction and Related Work}
In the age of information overload, timeline generation \cite{allan2001,chieu2004query,yan2011timeline,yan2011evolutionary}, through generating a series of concise summaries for a news event from a large collection of related documents, offers readers a fast way to grasp information. For the timeline generation of a long-term event, there exist two major difficulties. Firstly, the main topic of a long-term event is usually presented from several aspects or sub-topics. Each sub-topic can be regarded as a separate topic with its own evolving patterns. Secondly, these sub-topics are not independent. They usually exhibit strong correlations. This paper presents a hierarchical tree model to model the various sub-topics of one event and the dependencies among them.

Timelime generation was proposed by Allan et al.\cite{swan2000automatic} by extracting clusters of noun phrases and name entities. Later they built up a system to provide timelines which consist of one sentence per date by considering usefulness and novelty\cite{allan2001}. Chieu et al.\cite{chieu2004query} built a similar system in unit of sentences with interest and burstiness. However, these methods fail to explore the evolving characteristics of news events. Recently, Yan et al.\cite{yan2011timeline} extended the graph based sentence ranking algorithm used in traditional multi-document summarization (MDS) to timeline generation by projecting sentences from documents of different times into one plane. They further explored the timeline task through the optimization of a function by considering the combination of different respects such as relevance, coverage, coherence and diversity\cite{yan2011evolutionary,li2013evolutionary}. Time dependency is considered in Yan et al's work\cite{yan2011timeline,yan2011evolutionary}. Some other work uses graph based method \cite{wang2012generating,zhang2014patentline} to formulate timeline generation as a sentence ranking or optimization problem. But they fail to explore the topic information of the corpus or the structure of news information.

Topic models such as Latent Dirichlet Allocation (LDA)\cite{blei2003latent} and Hierarchical Dirichlet Processes (HDP)\cite{teh2006hierarchical} have been hot in these days for their capability in capturing latent topics within document collection. In topic models, documents are presented as a mixture distribution of topics and the latent topics can be aptly analyzed. Many revisions have been made to topic models to analyze topic distributions for document collection\cite{Ahmed2007,blei2006,caron2012generalized,ren2008dynamic,wang2006topics,zhang2010evolutionary,li2014timeline,li2014major,li2014weakly}.
However, existing topic models treat topics as paralleled and can not be used for exploring the hierarchical structure of topics in timeline generation.

Dirichlet Tree model, proposed by Dennis III\cite{dennis1991}, provides a proper structure for modeling hierarchical topic information. Such structure has been used for different purposes such as modeling word relations\cite{andrzejewski2009} or constructing vocabulary trees\cite{teh2009}. In standard Dirichlet Tree model, the structure of the tree has to be predefined. So it neither allows automatic learning of the tree structure nor accommodates the growing data. Another related work is the approach developed by Griffiths et al\cite{griffiths2004}. They built a hierarchical tree topic model with a non-parametric priors along with an effective inference algorithm call nested Chinese Restaurant Process (nCRP). The problem is, time and background information are not considered in their work. Our task is also different from novelty detection in TDT (Topic Detection and Tracking) task \cite{brants2003system,feng2007finding,fung2007time,zhang2007new}. TDT is a clustering task that finds clusters of documents matching real events by reducing the number of missing documents and the possibility of false alarms. What we are trying to do is to extract meaningful sentences that can help readers track the development of long-term events rather than find clusters of documents for novelty detection.

In this paper, we develop a novel topic model denoted as time-dependent Hierarchical Dirichlet Tree (tHDT) model to explore the hierarchical topic structure in the news corpus for timeline generation. Our model aptly combines Dirichlet Tree model with Dirichlet Processes and can automatically learn the structure of trees across corpus. Different levels of Markovian time dependency and background information are considered for tree structure construction. From that structure we can not only clearly see the evolution pattern of different aspects in the event, but also easily locate the topic position of each sentence. Based on the structure of the tree, we construct unified ranking framework by considering the relevance, coherence and coverage of each sentence. We build an experimental system on 8 real long-term events of public concern. The effectiveness of our model is verified through comparison with other baselines with regard to ROUGE matrix.
\section{DIRICHLET TREE}
\subsection{Dirichlet Tree}
\label{dp-hdp-sec}
The Dirichlet Tree distribution \cite{dennis1991} is a generalization of the Dirichlet distribution. It is a tree with words (or sentences) as leaf nodes. Let $\gamma^{(k)}$ denote the tree edges from the tree
root to node $k$, \emph{CH}(${k}$) denote the children of node $k$ and $L$ denote the leaves of the tree, $I$ denote the internal nodes and ${L(k)}$ denote the leaves in the tree whose root is node $k$. We can firstly draw a multinomial distribution at each internal node from $Dir(\gamma^{(k)})$ using weights from node $s$ to its children as the Dirichlet parameters. The probability $\phi^{(k)}$ of a word $k\in L$ is the product of the multinomial parameters on the edges from root to $k$. As is shown in Dennis's work \cite{dennis1991}, $p(\phi|\gamma)$ can be calculated in Equ.(2.1).
\begin{equation}
\begin{split}
p(\phi|\gamma)=&\Pi_{k\in L}[\phi^{k}]^{\gamma(k)-1}\\
&(\Pi_{s\in L}\frac{\Gamma(\sum^{C(s)}_{k}\gamma^{(k)})}{\Pi^{C(s)}_{k}\Gamma(\gamma^{(k)})}(\Sigma^{L(s)}_{k}\phi^{(k)})^{\Delta(s)}))
\end{split}
\end{equation}
where $\Delta(s) =\gamma^{(s)}-\sum^{C(s)}_{k}\gamma^{(k)}$ and $\Gamma(\cdot)$ denote the Gamma function. If $\Delta(s)=0$, Dirichlet Tree distribution would degenerate into a Dirichlet distribution. The Dirichlet Tree distribution is conjugate to the multinomial distribution and by integrating out $\phi$, we can easily calculate $p(\omega|\gamma)$ according to Equ.(2.2), where $w$ denotes a word and $n^{(k)}$ denotes the number of word $w$ appears in $L(k)$.
\begin{equation}
p(\omega|\gamma)=\Pi_{s\in L}(\frac{\Gamma(\sum^{C(s)}_k\gamma^{(k)})}{\Gamma(\sum^{C(s)}_{k}\gamma^{(k)}+n^{k})}\Pi^{C(s)}_{k}\frac{\Gamma(\gamma^{(k)}+n^{(k)})}{\Gamma(\gamma^{(k)})})
\end{equation}

\subsection{Dirichlet Processes(DP) and Hierarchical Dirichlet Processes(HDP)}
Dirichlet Process(DP) can be considered as a distribution over random measure $G$. We write $G\sim DP(\alpha,G_0)$ if\\
$(G(T_{1}), ..., G(T_{k})) \sim Dir(\alpha G_{0}(T_{1}), ..., \alpha G_{0}(T_{k}))$\\
$G_{0}$ is called base measure and $\alpha$ is concentration parameter. $G$ is a distribution over a given parameter space $\theta$, and we can draw parameter $\theta$ from $G$ as follows:
\begin{equation}
\theta_{n}|\theta_1,\theta_2,...,\theta_{n-1} \sim \sum_s\frac{m_s}{\sum_{l}m_{l}+\alpha}\delta(\phi_s)+\frac{\alpha}{\sum_{l}m_{l}+\alpha}G_0
\end{equation}
A HDP model defines a distribution over a set of DPs. In HDP, a global measure $G_{0}$ is distributed as a DP with concentration parameter $\alpha$ and base measure $H$. Then a set of local measures $G_j$ are drawn independently from $G_0$. $G_0 \sim DP(\alpha,H),G_j|G_0,\alpha \sim DP(\gamma,G_0)$\\
HDP model can be obtained from an equivalent Chinese Restaurant Franchise(CRF) metaphor \cite{teh2006hierarchical}. In CRF, each document is referred to a restaurant where words are compared to customers. Customers in the restaurant sit around different tables and each table is associated with a dish (topic) according to the dish menu. Let $n_i$ denotes the total number of customers having dinner in restaurant $i$. When a new customer $w_{ij}$ comes in, he can sit on the occupied table with probability $n_{iB}/(n_i-1+\gamma)$, sharing the dish (topic) $z_{iB}$ served at that table or picking an unoccupied table with probability $\gamma/(n_i-1+\gamma)$, where $n_{iB}$ denotes the number of customers already sitting around table $B$. The customer has to select a dish from the global dish menu if he chooses an unoccupied table. A dish that has already been shared in the global menu would be chosen with probability $m_k/(\sum_{k}m_k+\alpha)$ and a new dish with probability $\alpha/(\sum_{k}m_k+\alpha)$. Let $\theta$ be the parameter associated with table and $\phi$ be the parameter associated with the dish served at table. Let $G_j$ be the local restaurant dish menu and $G_0$ as the global dish menu, the Chinese Restaurant Process, denoted as $CRP(\alpha,\gamma,G_j,G_0)$, is illustrated in Equ.(2.4):
\begin{equation}
\begin{split}
&\theta_{i,j}|\theta_{i,1},...,\theta_{i,j-1},\alpha \\
&\sim \sum_{\phi=\theta_{ij}}\frac{n_{iB}}{n_{i}-1+\gamma}\delta_{\phi_{iB}}+\frac{\gamma}{n_{i}-1+\gamma}\delta_{\phi_{iB}^{new}}\\
&\phi_{jB}^{new}|\phi,\alpha\sim\sum_k\frac{m_k}{\sum_{i}m_{i}+\alpha}\delta_{\phi_k}+\frac{\alpha}{\sum_{i}m_{i}+\alpha}H\\
\end{split}
\end{equation}

\section{THDT MODEL FOR TIMELINE GENERATION}
\subsection{Problem Formulation}
Here we give a standard formulation of the task. Given a query, $Q=\{w_{qi}\}_{i=1}^{i=n}$, where $w_{qi}$ is the word in the query, we get a set of query related documents from the Internet.
The corpus is divided into a series of document collections according to the published time as $C=\{C^t\}_{t=1}^{t=T}$, where $C^t=\{D_{i}^t\}_{i=1}^{i=N^t}$, corresponding to the document collection published at time t. $D_i^t$ denotes the $i^{th}$ document at time $t$ and $N^t$ denotes the number of documents published at time t. Document $D_i^t$ is formulated as a collection of words $D_{i}^t=\{w_{in}^t\}_{n=1}^{n=N_i^t}$.
$V$ denotes the vocabulary size. The output of the algorithm is a series of timelines $I=\{I^t\}_{t=1}^{t=T}$ and $I^t\subset C^t$.
\subsection{tHDT} In this section, we would clarify the details of our tHDT model. To better illustrate our model, we firstly think about the generative process of how a writer usually works on a news report. When he decides to write a news article about Greek Crisis, he would firstly decide which aspect to write about. He may talk about particular aspects such as ``the evolution of crisis'', ``public protests'', ``the rescue package from EU'', or just write something about the background information. If he chooses ``the evolution of crisis'', he has to further decide the details he would like to talk about, maybe some background or some other aspects such as ``the danger of default'', ``Tax evasion and corruption'' or ``Greek's downgrading of creditworthiness''. At last, he has to further decide the vocabularies to use. If there are
lots of reports yesterday talking about the ``rescue package from EU'', it is more likely that the writer would track the development of this topic again. But of course, he may also change to a new aspect.

tHDT represents the distribution of topics in sentences by organizing topics into a tree of depth $L$. Each sentence is assigned to a path in a tree with a vector of topics. Each node in the tree is associated with a topic distribution over words. The sampling method alternates between choosing a new path for each sentence through the tree. The edge from internal node $k$ and $CH(k)$ is generated from a Hierarchical Dirichlet Process where both document and time information are considered. Since background information
has great influence on topic modeling, especially for the task of novelty detection \cite{chemudugunta2007}, we use latent parameter $y_{ijb}^{t(l)}$ to model different levels of background information for word $w_{ijb}^{t}$. $y_{ijb}^{t(l)}$ follows a binomial distribution $y_{ijb}^{t(l)}\sim binomial(\rho)$ with a beta prior $\rho\sim beta(\beta)$. For example, word ``Lehman'' should be regarded as an aspect word in the ``Financial Crisis'' but a background word in the subtopic ``Collapse of Lehman Brothers''.

Inspired by the work from Blei et al.\cite{blei2003latent}, we use a Chinese Restaurant Franchise Metaphor for Dirichlet Tree distribution, denoted as $DirTree(\alpha^1,\gamma^1,\alpha^2,\gamma^2,...,\alpha^{L-1},\gamma^{L})$. There are a series of restaurants in a city. Each restaurant has a series of tables with a dish on each table. Besides the dish, there is also a note specifying a list of dishes and the address of another restaurant on that table. For customer $w\in D_i^t\in C^t$, he will stay in the city for at most $L$ days. On the first day, he enters a restaurant and chooses a table and enjoys dish $z_{w}^{(1)}, z_{w}^{(1)}\sim CRP(\alpha^1,\gamma^1,G_{j}^{t(1)},G_{0}^{t(1)})$ according to Equ.(2.4). On the second day, he can choose to stay or leave. If he chooses to stay, he needs to go to the restaurant according to the address on the first day's table and select a dish from yesterday's dish list according to $z_w^{(2)}|z_w^{(1)}\sim CRP(\alpha^2,\gamma^2,G_{j}^{t(2)},G_{0}^{t(2)})$. The customer repeats this process before he chooses to leave for at most $L$ days and enjoys $l$ $(1\leq l\leq L)$ dishes $\{z_w^{(1)},z_w^{(2)},...,z_w^{(l)}\}$, which constitutes a path from the root of tree to the leaf.

To consider the time dependency in timeline generation, the global menu list at time t, $\{G_0^{t(l)}\}_{l=1}^{l=L}$ is drawn from a Dirichlet Process which combines of base measure $H$ and global
measure at $t-1$ with the Markovian assumption \cite{zhang2010evolutionary}:
\begin {equation}
G_0^{t(l)}\sim DP(\alpha^l,\nu(\delta_{t,t-1})G_0^{t-1,(l)}+(1-\nu(\delta_{t,t-1}))H)
\end {equation}\label{Equ:Mark}
where $\nu(\delta_{t,t-1})=exp(-\delta_{t,t-1}/\lambda)$, and it controls the influence of neighboring data. $H$ is the base measure. $\lambda$ is the decay factor of the time-decaying kernel. Similar approach can be found in many existing approaches \cite{zhang2010evolutionary,ren2008dynamic,Ahmed2007}. The generation process of tHDT model is shown in Figure 1.
\begin{figure}[!ht]
\rule{7.8cm}{0.03cm}
\vspace{-10pt}
\begin{tightitemize}
\item for each time $t\in [1:T]$
\begin{tightitemize}
\item for each level $l\in [1,L]$
~~~~~~~~~\item draw $G_0^{t(l)}$ according to Equ.(3.5)
\end{tightitemize}
\item for each document $D_i^t   i\in[1,N^t]$
\begin{tightitemize}
\item for each level $l\in [1,L]$
~~~~~~~~~\item draw $G_i^{t(l)}\sim DP(\gamma^t,G_0^{t(l)})$
\end{tightitemize}
\item for each word $w\in D_i^t\in C^t, i\in [1,N^t], t\in[1,T]$
\begin{tightitemize}
\item  draw path $\{z_w^{t(1)},z_w^{t(2)},...,z_w^{t(L)}\}\sim\\ DirTree(\alpha^1,\gamma^1,\alpha^2,\gamma^2,...,\alpha^L,\gamma^L)$
\item for $l \in [1,L]$:
~~~~~~~~~\item draw $\rho_w^l\sim beta(\beta)$
~~~~~~~~~\item draw $y_w^l\sim bionomial (\rho_w^l)$
~~~~~~~~~\item draw $w\sim w|z_w^{t(1)},...,z_w^{t(L)},y_w^{t(1)},...,y_w^{t(L)}$
\end{tightitemize}
\end{tightitemize}
\vspace{-0pt}
\rule{7.8cm}{0.03cm}
\caption{Generation Process of tHDT}\label{fig2}
\vspace{-0pt}
\end{figure}

\subsection{Inference}
For model inference, we use a straightforward Gibbs sampler based on the Chinese Restaurant Franchise. For customer $w\in D_i^t\in C^t$, given the dish he enjoys at $l^{th}$ day $z_w^{(l)}$, we would sample the dish $z_w^{(l+1)}$ he would enjoy on $(l+1)^{th}$ day according to CRF. Let $B_w^{(l+1)}$ denote the table that customer sits at on $(l+1)^{th}$ day and $z_B$ denote the dish served at table $B$. We alternate sample variables $y_w^{(l+1)},B_w^{(l+1)}$ and $z_w^{(l+1)}$ as follows:\\
\noindent {\bf sample $y_w^{(l)}$ for each word $w$:}
If $y_w^{(l+1)}=0$, $w$ is a background word for topic $z_w^{(l)}$. In CRF metaphor, a background customer will leave the city at $l^{th}$ day. If $y_w^{(l+1)}=0$, customer will stay in the city and choose a dish $z_w^{(l+1)}$ for $(l+1)^{th}$ day. $z_w^{(l+1)}$ is the child of $z_w^{(l)}$. Let $C_z^{(\cdot)}$ denote the total number of words that have been assigned to topic $z$ in current document. For simplication, we rewrite $z_w^{(l)}$ to $z$ and $z_w^{(l+1)}$ to $z'$. $C_z^{(0)}$ and $C_z^{(1)}$ are the number of words assigned to background and aspect topic. $C_{z'}$ is the number of words assigned to topic $z'$. We have $C_z^{(1)}=\sum_{z'\in CH(z)}C_{z'}$. $\eta$ is a hyperparameter. $CH(z)$ denotes the children of $z$ in the tree structure and $K=|CH(z)|$. We have:
\begin{equation}
\begin{aligned}
&P(y_w^{(l)}|y/y_w^{(l)},z_s^{(l)})\propto \\
&\left\{
\begin{aligned}
&\frac{C_z^{(0)}+\beta}{C_z^{(\cdot)}+2\beta} \frac{E_{z(0)}^w+\lambda}{\sum_{(w')}E_{z(0)}^{w'}+V_z\lambda}, if ~y_w^{t(l)}~=~0\\
&\frac{C_z^{(1)}+\beta}{C_z^{(\cdot)}+2\beta}\sum_{z'\in CH(z)}\frac{C_{z'}+\eta}{C_z^{(1)}+K\eta}\frac{E_{z'}^w+\lambda}{\sum_{(w')}E_{z'}^{w'}+V_z\lambda}, if ~y_w^{t(l)}~=~1
\end{aligned}
\right.
\end{aligned}
\end{equation}

\noindent {\bf sample table $B$ for current word $w$:}
\begin{equation}
\begin{aligned}
&P(B_w=B|w,z_s^{(l)},z_B=z')\propto\\
&~~~~~~~~~~~~~~~~~\left\{
\begin{aligned}
&n_{B}\frac{E_{z'}^w+\lambda}{\sum_{w'}E_{z'}^{w'}+V_z\lambda} &if ~B~is~used\\
&\gamma^l/V_z &if~B~is~new
\end{aligned}
\right.
\end{aligned}
\end{equation}
\noindent {\bf sample dish $z_B$ for the new table: }
\noindent If the customer chooses a new table, we have to sample a dish for this table.
\begin{equation}
\begin{aligned}
&P(z_{B_{new}}=z'|w\in s,z)\propto\\
&~~~~~~~~~~~~~~\left\{
\begin{aligned}
&m_{tz'}^{'}\frac{E_{z'}^w+\lambda}{\sum_{w'}E_{z'}^{w'}+V_z\lambda} &if~z'~is~used\\
&\alpha^l/V_z &if~z'~is~new
\end{aligned}
\right.
\end{aligned}
\end{equation}
where $m_{tk}$ denotes the number of tables having dish $k$ at time $t$. According to Equ.(3.5), by incorporating Markovian assumption, we have $m_{tk}^{'}=v(\delta_{t,t-1})m_{t-1,k}+(1-v(\delta_{t,t-1}))m_{t,k}$.

\noindent {\bf re-sample dish $z_B$ for each table: }
\noindent Since the process of dish sampling actually changes the component member of tables after each iteration, we need to re-sample dish for each table. $R_B^{(l)}$ denote the collection of customers sitting around table B at $l^{th}$ level, $R_B^{(l)}=\{w|w\in D_i^t,B_w=B\}$ and $N_{R_B^{(l)}}^w$ denote the number of replicates of word $w$ at table $B$. We have:
\begin{equation}
\begin{aligned}
&P(z_B=z'|D_i^t,y_w)\propto\\
&\left\{
\begin{aligned}
&\frac{m_{tk}^{'}\Gamma(\sum_{w'}E_{z'}^{w'}+V_z\lambda)}{\Gamma(\sum_{w'}E_{z'}^{w'}+|R_B^{(l)}|+V_z\lambda)}\Pi_{w\in R_B^{(l)}}\\
&\frac{\Gamma(E_{z'}^{w}+N_{R_B^{(l)}}^w+\lambda)}{\Gamma(E_{z'}^{w}+\lambda)} &if~z'~is~used\\
&\alpha^l\frac{\Gamma(V_z\lambda)}{\Gamma(|R_B^{(l)}|+V_z\lambda)}\Pi_{w\in R_B^{(l)}}\frac{\Gamma(N_{R_B^{(l)}}^w+\lambda)}{\Gamma(\lambda)} &if~z'~is~new
\end{aligned}
\right.
\end{aligned}
\end{equation}
After Gibbs sampling, we can calculate the probability that a word is generated by a path as follows:
\begin{equation}
\begin{aligned}
P(w|z=\{z^1,z^2,...,z^L\})=\prod_{l=1}^L\frac{C_{zl}}{C_{zl-1}}\frac{E_{zl}^w+\lambda}{\sum_{w'}E_{zl}^{w'}+V_z\lambda}
\end{aligned}
\end{equation}

\subsection{Tree-based Sentence Selection}
tHDT model constructs a hierarchical structure by locating candidate words on the nodes of tree. Each word is represented by a path and we assume that words sharing similar paths should be more similar to each other as they share the similar topics. The similarity between two words $w_1$ and $w_2$ is obtained by firstly calculating the Jensen-Shannon divergence between two sentences\footnote{Since Kullback-Leibler divergence is asymmetric, we adopt  Jensen-Shannon divergence in this paper, which is symmetric and nonnegative.}.
$$JS(w_1||w_2)=KL(w_1||\frac{w_1+w_2}{2})+KL(w_2||\frac{w_1+w_2}{2})$$
$$KL(w_1||w_2)=\sum_{z_1,...,z_{L-1}}P(w_1|z_1,...,z_L)log\frac{P(w_1|z_1,...,z_L)}{P(w_2|z_1,...,z_L)}$$
Let $G_i$ denote a collection of words. $G_i=\{w|w\in G_i\}$. $q(G|z_1,...,z_L)=\frac{1}{|G|}\sum_{w\in G}p(w|z_1,...,z_L)$. And the KL divergence between two collections of words are defined as follows:
$$KL(G_1||G_2)=
\sum_{z_1,...,z_L}q(G_1|z_1,...,z_L)log\frac{q(G_1|z_1,...,z_L)}{q(G_2|z_1,...,z_L)}$$

\noindent JS divergence is then transformed into a similarity measure by an decreasing logistic function $1/(1+e^x)$\cite{yan2011evolutionary} to refine the relevance in the range of (0,1).

Let $s$ denote a sentence comprised of a collection of words. A good timeline should properly consider the following key requirements\cite{li2009enhancing}:\par
\noindent(1){\bf Focus}: The timeline for epoch $t$ should be related to the given query. Since the Query distribution is too sparse, we adopt the strategy taken in Yan et al.'s work\cite{yan2011evolutionary}. Query expansion is introduced by pseudo-relevance
feedback to enlarge Q. We retrieve $top-\kappa$ snippets, which is denoted as $Q'$ and use $Q'$ to approximate $Q$.
\begin{equation*}
\zeta_F(s)=\frac{1}{1+exp[JS(s\|Q')]}
\end{equation*}

\noindent(2){\bf Coherence}: A timeline consists of a series of individual but correlated sentences. News evolves over time and a good timeline should be coherent with neighboring documents so that the timeline can track the evolution pattern of news.
\begin{equation*}
\zeta_{CH}(s)=\frac{1}{1+exp[JS(s\|C_{t-1})]}
\end{equation*}
\noindent (3){\bf Coverage}:
Document $D_i^t$ should keep alignment with source documents at epoch $t$ and the selected sentence should conclude the important aspects mentioned in $D_i^t$.
\begin{equation*}
\zeta_{CV}(s)=\frac{1}{1+exp[JS(s\|D_s)]}\cdot\frac{1}{1+exp[JS(D_s\|C_t)]}
\end{equation*}
\noindent Given the source collection, each sentence is scored based on the weighted combination of these three requirements. The score function is illustrated as follows:
\begin{equation}
\begin{aligned}
Score(s)=w_1\cdot\zeta_{F}(s)+w_2\cdot\zeta_{CH}(s)+w_3\cdot\zeta_{CV}(s)
\end{aligned}
\end{equation}

\noindent $\sum_iw_i=1$. Sentences with higher scores are selected. To avoid aspect redundancy, MMR strategy\cite{goldstein1999summarizing} is adopted in the process of sentence selection.

\section{EXPERIMENTS}
\subsection{Datasets and Experiments Setup}
\begin{table}
\centering
\begin{tabular}{c|c|c|c} \hline
News Sources&Nation&News Sources&Nation\\\hline
ABC&US&Wanshington Post&US\\\hline
CNN&US&New York Times&US\\\hline
BBC&UK&Xinhua&China\\\hline
\end{tabular}
\caption{News Sources for timeline generation}
\end{table}
There is no existing standard evaluation datasets for timeline generation. In this paper, we build 8 datasets about real long-term events of public concern and use the golden standards to evaluate the performance of different models. 
We download 9935 news articles from different resources and the details are illustrated at Table1 and Table2. Dataset 1 ``Iraq War", 2 ``Apple Inc", 6 ``Afghanistan War" and 8 ``North Korea's Nuclear Crisis" are used as training sets for parameter tuning and the rest are used for testing.
\begin{table}
\hspace{-5pt}
\begin{tabular}{l|l|l|l|l} \hline
Topics&\#Docs&\#Epochs&From&To\\\hline
Iraq War&1071&318&2003&2005\\\hline
Apple Inc&716&217&2007&2011\\\hline
Financial Crisis&860&254&2007&2008\\\hline
Greek Debt Crisis&1031&372&2010&2012\\\hline
Arab Spring&1089&232&2010&2012\\\hline
Afghanistan War&1870&487&2001&2005\\\hline
US Election&1246&286&2011&2012 \\\hline
North Korea's&1078&266&2007&2012\\
Nuclear Crisis&&&&\\\hline
\end{tabular}
\caption{News Sources for timeline generation}
\vspace{-10pt}
\end{table}
\subsection{Evaluation Metrics}
We adopt ROUGE toolkit (version 1.5.5) for performance evaluation. The timeline quality is measured
by counting the number of overlapping units, such as N-gram, word sequences and word pairs between candidate timeline CT and the ground-truth timeline GT. Several automatic evaluation methods are implemented in ROUGE and each of the methods can generate scores of recall, precision and F-measure.
Reference timelines in ROUGE evaluation are manually generated by using Amazon Mechanical Turk\footnote{http://mturk.com}. Workers were asked to generate reference timelines for news at each
epoch in 30 words and we collect 2432 timelines in total.
\subsection{Parameter Tuning}
To tune the parameters of $w_i(i = 1,2,3)$, $\lambda$ and $L$ in our system, we adopt a gradient search strategy. Keeping other parameters fixed, we vary one parameter at a time to examine the changes of its performance on training data.
We firstly set the number of tree level $L$ to $2$, $w_i(i = 1,2,3)$ to $1/3$ and perform experiments on setting different values of $\lambda$, ranging from $1$ to $20$ with interval of $1.0$. ROUGE scores reach its peak at round $12$ and drops afterwards. Then we fix the value of $\lambda$ to $12$ and tune $w_i(i = 1,2,3)$ in the greedy way. We gradually change the value of $w_1$ from $0$ to $1$ with interval of $0.05$, with simultaneously fixing both $w_2$ and $w_3$ to the values of $(1-w_1)/2$. The performance gets better as $w_1$ increases from $0$ to $0.20$ and then declines gently. Next we set the value of $w_1$ to $0.20$ and change the value of $w_2$ from $0$ to $0.80$ with interval of $0.05$. According to experimental results, the value of $w_2$ is set to $0.45$, and $w_3$ set to $0.35$ correspondingly. At last, we perform experiments setting values of $L$ ranging from $1$ to $5$. Specifically, when $L=1$, the model degenerates into time-dependent HDP model. Experiments prove that we get better scores when $L$ is set to $2$ and $3$. This is partly because a $2$ or $3$ layer tree may better approximate the hierarchical structure of topics in news.


\subsection{Performance Comparison with Baselines}
We implement the following algorithms as the baseline systems. Some are designed for traditional summarization without temporal dimension. For fairness we conduct the same preprocessing for all algorithms and truncate the timeline for each day to 30 words. \\
{\bf Random}: The methods that select sentences randomly for timeline generation.\\
{\bf Centroid}: The method that applies MEAD algorithm \cite{radev2004centroid}, which has been widely used in MDS for sentence selection according to centroid value, positional value, and first-sentence overlap. \\
{\bf GMDS}: The Graph-based MDS method proposed by Wan et al.\cite{wan2008multi}
that constructs a sentence connectivity graph based on
cosine similarity and then selects important sentences based on centrality.\\
We also have the system of Chieu et al.\cite{chieu2004query} and Yan et al.\cite{yan2011evolutionary} for comparison.

\begin{figure}[!ht]
\centering
  \includegraphics[width=2.8in]{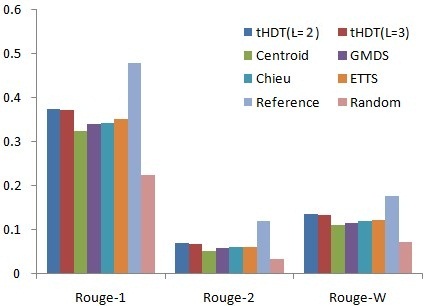}
  \caption{Overall Performance of Different Baselines.}\label{overall_1}
  \vspace{-10pt}
\end{figure}
\begin{table}[!ht]
\centering
{\small
\begin{tabular}{|c|c|c|c|c|c|c|} \hline
&\multicolumn{3}{|c|}{Financial Crisis}&\multicolumn{3}{|c|}{Greek Debt Crisis}\\\hline
systems&R-1&R-2&R-W&R-1&R-2&R-W\\\hline
(L=2)&.382&\textbf{.079}&.142&.366&\textbf{.064}&\textbf{.136}\\\hline
(L=3)&\textbf{.386}&.078&\textbf{.144}&\textbf{.370}&.062&\textbf{.136}\\\hline
Centroid&.297&.039&.081&.279&.031&.087\\\hline
GMDS&.286&.044&.088&.288&.033&.081\\\hline
Chieu&.335&.056&.115&.313&.048&.119\\\hline
ETTS&.340&.054&.120&.327&.044&.120\\\hline
Random&.227&.031&.071&.213&.028&.068\\\hline\hline
&\multicolumn{3}{|c|}{American Election}&\multicolumn{3}{|c|}{Arab Spring}\\\hline
systems&R-1&R-2&R-W&R-1&R-2&R-W\\\hline
(L=2)&\textbf{.384}&\textbf{.074}&\textbf{.132}&\textbf{.446}&\textbf{.096}&\textbf{.165}\\\hline
(L=3)&.378&.072&.128&.433&.090&.157\\\hline
Centriod&.285&.038&.094&.332&.048&.098\\\hline
GMS&.306&.044&.095&.341&.054&.109\\\hline
Chieu&.353&.058&.112&.366&.076&.124\\\hline
ETTS&.359&.062&.117&.379&.075&.132\\\hline
Random&.224&.027&.070&.230&.035&.076\\\hline

\end{tabular}
}
\caption{Comparison with different baselines}
\hspace{-13pt}
\vspace{-13pt}
\end{table}

We average F-score performance in terms of ROUGE-1, ROUGE-2, and ROUGE-W on all sets. The overall results are shown in Figure \ref{overall_1} and the details are shown in Table 3. As we can see, since traditional MDS methods such as {\bf Centriod} and {\bf GMDS} only consider sentence ranking and selection strategy and ignore the evolution pattern of topics. Many sentences
are missed, resulting in a low recall. {\bf Chieu} does not capture time attributes, so its results are worse than {\bf ETTS}, which uses local and global measures to capture the time dependencies. However, as {\bf ETTS} is a sentence ranking algorithm in nature, its sentence selection strategy may be biased for neglecting the evolution pattern of topic information across the corpus. {\bf tHDT} achieves the best results for its capability in detecting the hierarchical structures in the corpus. {\bf tHDT(L=2)} outperforms {\bf ETTS} by $6.2\%$, $12.9\%$ and $11.7\%$ with regard to the overall performance in ROUGE-1, ROUGE-2 and ROUGE-W respectively.
\subsection{Comparison with Other Topic Models}
To illustrate the effectiveness of our topic model, we provide six other baseline systems which adopt different modeling techniques.\par
{\bf tHDT-noBG}: A simplified version of tHDT without considering background information.\par
{\bf HDT}: Also a simplified version of tHDT with temporal information neglected\footnote{The value of $L$ is set to 2 in tHDT-noBG and HDT.}.\par
{\bf t-HDP}: A time-dependent HDP model without considering hierarchical structure of topics. It is a simple version of tHDT where $L=1$.\par
{\bf D-LDA}: A dynamic LDA\cite{blei2006} where topic-word distribution and popularity are linked across epochs by including Markovian assumption\footnote{Topic number is set to 80 in D-LDA and Stand-LDA.}.\par
\begin{figure}[!ht]
\centering
  \includegraphics[width=2.8in]{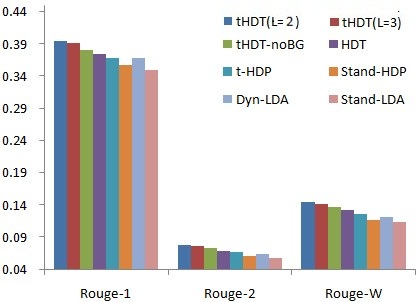}
  \caption{Overall Performance of Different Models}\label{topic}
  \vspace{-10pt}
\end{figure}

We also use the standard LDA and HDP model as baselines.\\
The overall results are shown in Figure \ref{topic} and details are listed in Table 4.
As we can see, {\bf tHDT}(L=2,3) is better than {\bf tHDT-noBG}, which verifies that the identification of background words can promote the performance of timeline generation.
{\bf tHDT} is also better than {\bf HDT} which verifies the necessity of modeling temporal information in topic modeling for timelines. 
We can also see that {\bf tHDT} achieves better results than {\bf t-HDP} and {\bf D-LDA} as the latter two models treat different topics as paralleled and do not explore the hierarchical structure of topic information. Compared with Table
3 and Figure 2, we find that most topic based models can get better results than baselines. This illustrates the advantages of topic modeling in timeline generation over methods that only focus on sentence selection strategy.
\begin{table}[!ht]
{\small
\begin{tabular}{|c|c|c|c|c|c|c|} \hline
&\multicolumn{3}{|c|}{Financial Crisis}&\multicolumn{3}{|c|}{Greek Debt Crisis}\\\hline
systems&R-1&R-2&R-W&R-1&R-2&R-W\\\hline
(L=2)&.382&{\bf .079}&.142&.366&{\bf .064}&{\bf .136}\\\hline
(L=3)&{\bf .386}&.078&{\bf .144}&{\bf .370}&.062&{\bf .136}\\\hline
(noBG)&.373&.074&.132&.342&.058&.130\\\hline
HDT&.362&.065&.131&.337&.052&.122\\\hline
t-HDP&.360&.064&.128&.338&.053&.113\\\hline
HDP&.348&.055&.114&.323&.050&.108\\\hline
D-LDA&.358&.060&.120&.330&.048&.111\\\hline
LDA&.334&.046&.110&.319&.050&.107\\\hline\hline
&\multicolumn{3}{|c|}{American Election}&\multicolumn{3}{|c|}{Arab Spring}\\\hline
systems&R-1&R-2&R-W&R-1&R-2&R-W\\\hline
(L=2)&{\bf .384}&{\bf .074}&{\bf .132}&{\bf .446}&{\bf .096}&{\bf .165}\\\hline
(L=3)&.378&.072&.128&.433&.090&.158\\\hline
(noBG)&.376&.070&.128&.431&.089&.159\\\hline
HDT&.373&.068&.124&.426&.086&.151\\\hline
t-HDP&.366&.068&.121&.410&.080&.138\\\hline
HDP&.361&.062&.114&.398&.073&.132\\\hline
D-LDA&.368&.070&.120&.413&.078&.135\\\hline
LDA&.352&.062&.110&.392&.072&.126\\\hline
\end{tabular}
}
\caption{Comparison with Other Topic Models}
\vspace{-20pt}
\hspace{-13pt}
\end{table}
\section{Conclusion}
We develop a novel topic model denoted as time-dependent Hierarchical Dirichlet Tree (tHDT) to explore the hierarchical topic structure for timeline generation. Our model aptly combines Dirichlet Tree with Dirichlet Processes and can automatically learn the structure of trees across corpus. Different levels of Markovian time dependency and background information are considered for tree structure construction. We build an experimental system on 8 real long-term events of public concern. Experimental results illustrate the effectiveness of our proposed model.

\bibliographystyle{}
\bibliography{sigproc}
\newpage
~\bf{Appendix}
\begin{figure*}[!ht]
\centering
  \includegraphics[width=5.0in]{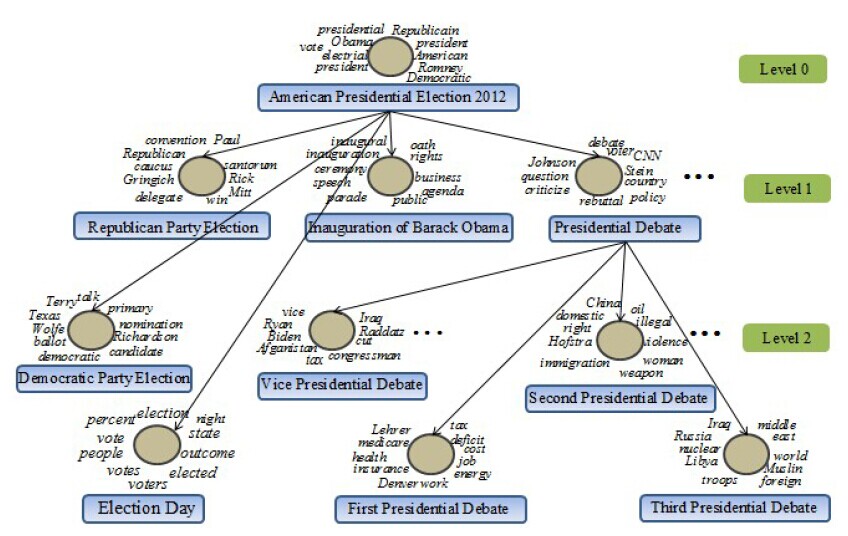}
  \caption{A sample of two-level hierarchical structure in American Presidential Election 2012. Each node denotes a topic and surrounding words denote the top words in that topic. The title for each topic is manually labelled.}\label{fig.1001}
\end{figure*}
\begin{figure*}[!ht]
\centering
  \includegraphics[width=5.0in]{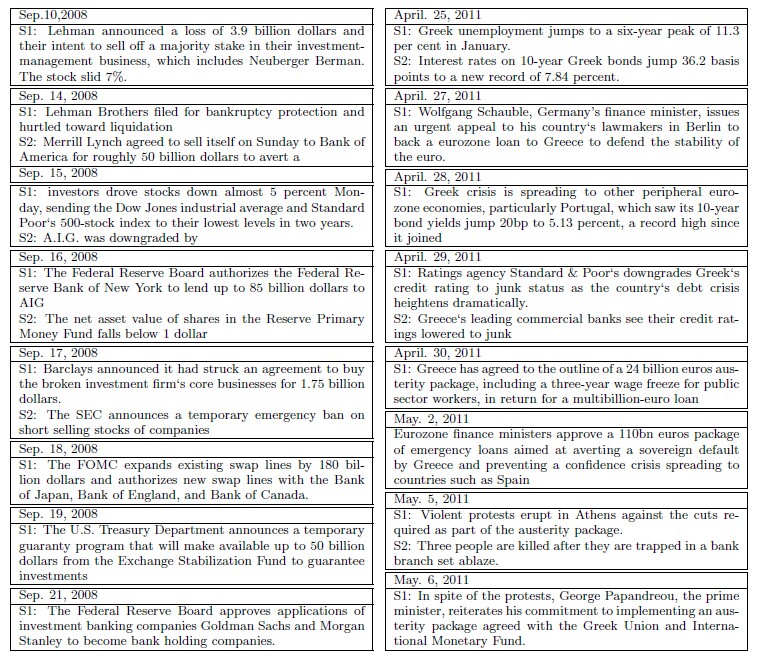}
  \caption{Selected part of timeline generated by tHDT for (a) Financial crisis and (b) Greek Debt Crisis. For each day, timeline is truncated to 30 words.}\label{fig.1002} 
\end{figure*}

\end{document}